\documentclass[10pt,twocolumn,letterpaper]{article}

\usepackage[pagenumbers]{iccv} 

\usepackage{multirow}

\definecolor{iccvblue}{rgb}{0.21,0.49,0.74}
\usepackage[pagebackref,breaklinks,colorlinks,allcolors=iccvblue]{hyperref}

\usepackage{accents}

\newcommand{\notsosmall}{\fontsize{10pt}{12pt}\selectfont}

\definecolor{somegray}{rgb}{0.5, 0.5, 0.5}
\newcommand{\darkgrayed}[1]{\textcolor{somegray}{#1}}
\makeatletter
\newcommand*\titleheader[1]{\gdef\@titleheader{#1}}
\AtBeginDocument{
  \let\st@red@title\@title
  \def\@title{
    \vskip-3em
    \bgroup\normalfont\large\centering\@titleheader\par\egroup
    \vskip1.5em\st@red@title}
}
\titleheader{\darkgrayed{This paper has been accepted for publication at the \\
IEEE/CVF International Conference on Computer Vision (ICCV), Honolulu, 2025.
\copyright IEEE}}

\title{
Depth AnyEvent: A Cross-Modal Distillation Paradigm for Event-Based Monocular Depth Estimation
}

\author{Luca Bartolomei$^{*,\dagger}$ \hspace{0.7cm} Enrico Mannocci$^{\dagger}$ \hspace{0.7cm} Fabio Tosi$^\dagger$ \hspace{0.7cm} Matteo Poggi$^{*,\dagger}$  \hspace{0.7cm} Stefano Mattoccia$^{*,\dagger}$ \\
\notsosmall $^*$Advanced Research Center on Electronic System (ARCES) \\ 
\notsosmall $^\dagger$Department of Computer Science and Engineering (DISI) \vspace{-0.1cm}\\
\notsosmall University of Bologna, Italy \\
{\tt\small\{luca.bartolomei5, fabio.tosi5, m.poggi, stefano.mattoccia\}@unibo.it} \\
\small\url{https://bartn8.github.io/depthanyevent}
}

\begin{document}

\twocolumn[{
    \renewcommand\twocolumn[1][]{#1}
    \maketitle
    \centering
    \vspace{-0.5cm}
    \includegraphics[width=0.95\textwidth]{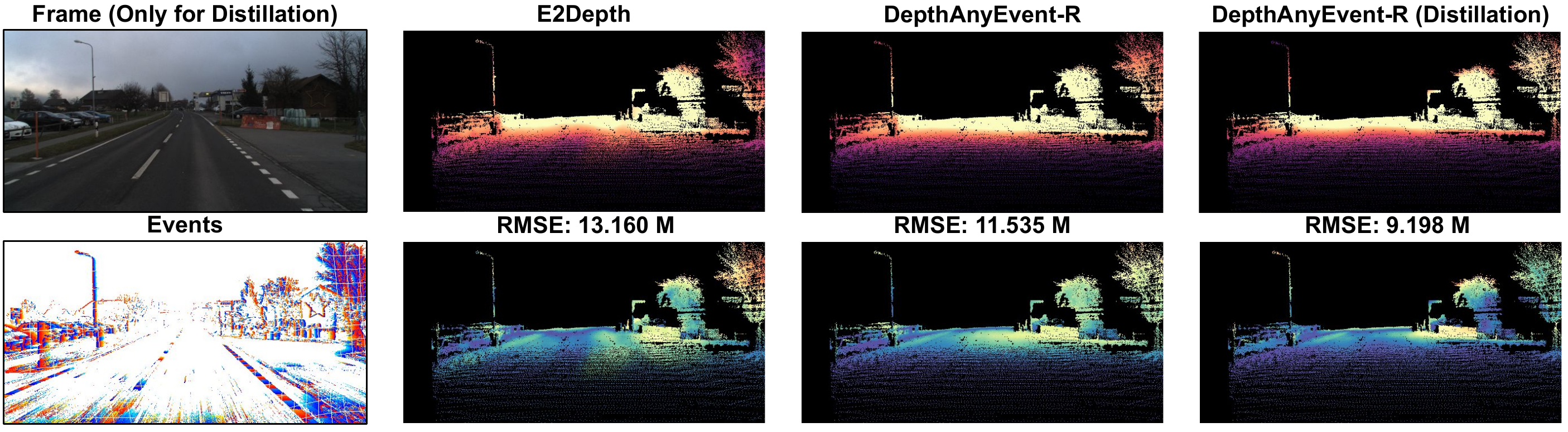}\vspace{-0.3cm}
    \captionof{figure}{\textbf{DepthAnyEvent-R in action.} The first column shows the input frame (used only for distillation) and the corresponding event visualization. The other three columns present depth estimation results from different approaches: E2Depth \cite{hidalgo2020learning}, our DepthAnyEvent-R, and our DepthAnyEvent-R trained with our distillation approach. The top row shows the estimated depth maps while the bottom row depicts their corresponding RMSE visualizations.}\vspace{0.3cm}
    \label{fig:teaser}
}]

\begin{abstract}

Event cameras capture sparse, high-temporal-resolution visual information, making them particularly suitable for challenging environments with high-speed motion and strongly varying lighting conditions. However, the lack of large datasets with dense ground-truth depth annotations hinders learning-based monocular depth estimation from event data. To address this limitation, we propose a cross-modal distillation paradigm to generate dense proxy labels leveraging a Vision Foundation Model (VFM). 
Our strategy requires an event stream spatially aligned with RGB frames, a simple setup even available off-the-shelf, and exploits the robustness of large-scale VFMs.
Additionally, we propose to adapt VFMs, either a vanilla one like Depth Anything v2 (DAv2), or deriving from it a novel recurrent architecture to infer depth from monocular event cameras. 
We evaluate our approach with synthetic and real-world datasets, demonstrating that i) our cross-modal paradigm achieves competitive performance compared to fully supervised methods without requiring expensive depth annotations, and ii) our VFM-based models achieve state-of-the-art performance.

\end{abstract}    
\section{Introduction}
\label{sec:intro}

Depth perception from cameras is paramount for many application fields, such as those concerning the autonomous navigation of agents in complex scenarios or robotic tasks. In these fields, learning-based methods using conventional cameras have obtained compelling results in the last decade. Moreover, this paradigm enabled inferring depth from a single camera, which brings significant advantages compared to multicamera setups in terms of cost, calibration complexity, and physical constraints.
Nonetheless, conventional camera systems struggle to provide a prompt and reliable perception of the sensed environment when dealing with highly dynamic scenes resulting from the fast movement of vehicles, drones, robots or in the presence of challenging illumination conditions such as high contrast scenarios, low light, or rapid lighting changes. These limitations are intrinsic to the conventional camera acquisition technology occurring at discrete periodic intervals and with a limited dynamic range, causing motion blur, over/under exposure, and potentially missing critical information between frames.
In contrast, the intrinsic ability to capture scene changes as soon as they appear -- with microsecond temporal resolution -- and the much higher dynamic range made event cameras \cite{EVENT_SURVEY} ideal for coping with the challenging application fields mentioned above. Event cameras only register brightness changes at each pixel independently, offering exceptional temporal resolution and robustness to lighting variations. However, these features come at the cost of meager information content compared to conventional cameras. Event cameras provide meaningful cues only for a small subset of the framed image with sufficient texture to trigger events, making depth perception from these devices extremely challenging. Moreover, the lack of large datasets with dense ground truth annotations further exacerbates this inherent difficulty, as collecting precise depth ground truth for event data remains costly and technically demanding.

To tackle these issues in a monocular event camera setup, we propose to leverage the effectiveness of image-based Vision Foundation Models (VFMs) for monocular depth estimation.  They have demonstrated remarkable capabilities through extensive pretraining on vast image collections, enabling robust depth prediction even in challenging scenario. As the first contribution, given sequences of aligned images and events, we propose a cross-modal distillation strategy that allows us to obtain dense proxy labels from a VFM to train event-based networks. This approach effectively transfers knowledge from the data-rich image domain to the data-sparse event domain. For our purposes, an off-the-shelf device like a DAVIS Camera \cite{DAVIS_CAMERA,DAVIS_SCARAMUZZA} that incorporates a conventional global shutter camera and an event-based sensor in the same pixel array would suffice to gather spatially aligned event streams and RGB frames.

Additionally, as the second contribution, we propose to adapt VFMs for event-based monocular depth estimation, either using a vanilla model like Depth Anything v2 (DAv2) or a novel recurrent architecture derived from it. 
To prove the effectiveness of our proposals, we assess the performance with synthetic and real-world datasets, showing that our cross-modal distillation paradigm allows for achieving competitive performance compared to fully supervised approaches, disregarding the need for expensive depth annotation. Moreover, adapting VFMs for monocular depth estimation according to our two proposals is state-of-the-art, setting new benchmarks for event-based depth estimation.

Figure \ref{fig:teaser} shows the compelling performance of our proposals, and our contributions can be summarized as follows:

\begin{itemize}
    \item A novel cross-modal distillation paradigm that leverages the robust proxy labels obtained from image-based VFMs for monocular depth estimation. 
    \item An adapting strategy to cast existing image-based VFMs into the event domain effortlessly.
    \item A novel recurrent architecture based on an adapted image-based VFM.
    \item Adapting VFMs to the event domain yields state-of-the-art performance, and our distillation paradigm is competitive against the supervision from depth sensors.      
\end{itemize}    

\section{Related Work}
\label{sec:related_work}
\textbf{Image-Based Monocular Depth Estimation.} Monocular depth estimation has evolved from traditional approaches \cite{4531745} to deep learning methods \cite{eigen2014depth,laina2016deeper}. Self-supervised techniques\cite{godard2017unsupervised,zhou2017unsupervised,DBLP:journals/corr/abs-1806-01260} have emerged to address this challenge of limited ground truth data by recasting depth estimation as an image reconstruction task using stereo images or videos. These approaches have been particularly valuable where dense depth annotations are expensive to obtain.  A significant step came with affine-invariant models \cite{Ranftl2022, Ranftl2021} that estimate depth up to an unknown scale and shift, allowing impressive cross-domain generalization capabilities. MiDaS~\cite{Ranftl2022} pioneered this direction by training on diverse large-scale datasets, followed by DPT~\cite{Ranftl2021} and more recently, the Depth Anything series~\cite{depthanything, depth_anything_v2}. These latter models represent the first generation of Visual Foundation Models for monocular depth estimation, leveraging large-scale pretraining and diverse data sources to achieve unprecedented robustness. The effectiveness of these models lies in their ability to combine knowledge from various domains, including internet photo collections~\cite{li2018megadepth, Yin2020}, LiDAR from autonomous driving scenarios~\cite{Geiger2012CVPR}, and RGB-D sensors~\cite{Silberman2012}. 
Recent advances in VFMs have focused on improving metric accuracy through camera parameter integration~\cite{Yin2023, Guizilini2023}, leveraging generative approaches like diffusion models~\cite{Ji2023, Duan2023, Saxena2023}, and addressing temporal consistency\cite{shao2024learning}. Furthermore, attention-based architectures and transformer models ~\cite{zhao2022monovit} have shown significant improvements in capturing long-range dependencies crucial for accurate depth.  Despite recent advances, applying these methods to event-based cameras is still limited by the lack of large-scale annotated datasets. We tackle this by distilling knowledge from frame-based VFMs, enabling accurate depth estimation without costly event data annotations.

\textbf{Event-based Monocular Depth Estimation.} Event-based depth estimation began with supervised approach using recurrent architectures ~\cite{hidalgo2020learning, gehrig2021combining, liu2024event} designed to process the temporal information contained in event streams. Advanced models like~\cite{gehrig2021combining} further expanded this concept by fusing event and RGB data to exploit their complementary charactetistics. Multimodal fusion techniques have also been explored, combining events with LiDAR to generate dense depth maps \cite{9686362}. To address the scarcity of labeled event data, self-supervised methods have emerged as promising alternatives. Zhu et al.~\cite{zhu2019unsupervised} developed a framework that jointly estimates depth, optical flow, and camera poses using stereo consistency and motion blur minimization as training signals. Subsequent work~\cite{zhu2024selfsupervisedeventbasedmonoculardepth} eliminated the need for stereo setups by leveraging pose information from consecutive RGB frames aligned with the event camera, enabling dense depth estimation. Despite these advances, event-based depth estimation still falls short compared to frame-based methods.

\section{Preliminaries: Event Depth Estimation}
\label{sec:pre_method}

Event cameras measure the logarithmic change in brightness over time, and when it changes over a threshold $\pm C$, the associate pixel at position $(x_k,y_k)$ emits at time $t_k$ an asynchronous signal $e_k=(x_k,y_k,p_k,t_k)$ called \textit{event}.
Depending on the sign of this change, the event will have polarity $p_k\in\{-1,1\}$.
Each pixel of the $W \times H$ sensor grid of the event camera can independently emit events at any time, producing an asynchronous stream of events $\mathcal{E}=\{e_k\}^N_{k=1}$, where $N$ is the total number of fired events.

Given the event history $\mathcal{E}$, previous event-based dense monocular depth estimation models \cite{hidalgo2020learning,liu2024event,gehrig2021combining} convert the flow of events into a $\mathbf{E} \in \mathbb{R}^ {W \times H \times C}$ structured representation -- such as Voxel Grids \cite{zhu2019unsupervised} -- since the sparse structure of $\mathcal{E}$ is not suitable for standard CNNs. 
Intentionally, to estimate a depth map $\mathbf{D} \in \mathbb{R}^{W \times H}$ at a given timestamp $t_d$, events are retrospectively sampled from the stream $\mathcal{E}$, either within a fixed time window (SBT) -- \ie, $\mathcal{E}^{\Delta T}_{t_d} = \left\{ e_k \in \mathcal{E} \mid t_d - \Delta T \leq t_k \leq t_d \right\}$ -- or up to a predefined number $K$ of events (SBN) -- \ie, $\mathcal{E}^K_{t_d} = \left\{ e_k \in \mathcal{E} \mid d-K \leq k \leq d \right\}$ -- and subsequently \textit{stacked} using different strategies, including:

\textbf{Voxel Grid} \cite{zhu2019unsupervised}: The time interval used for sampling events is divided into $B$ uniform bins, where event polarities are accumulated using linear interpolation within each bin of a $\mathbf{E} \in \mathbb{R}^ {W \times H \times B}$ stack.

\textbf{Image-like} \cite{liu2024event}: A color-based representation where the R and B channels encode positive and negative polarities, respectively, resulting in an RGB image, \ie a $\mathbf{E} \in \mathbb{R}^ {W \times H \times 3}$ stack. Unlike the Voxel Grid representation, it does not retain temporal information.

\textbf{Tencode} \cite{Huang_2023_WACV}. A color image representation in which R and B channels encode positive and negative polarities, with G encoding the timestamp relative to the total time-lapse. It produces an RGB image, \ie a $\mathbf{E} \in \mathbb{R}^ {W \times H \times 3}$ stack.

For the sake of space, we report only the event representations relevant to our work, but additional details regarding event representations can be found in \cite{bartolomei2024lidar,ghosh2024event}.

\section{Proposed Method}
\label{sec:method}

\begin{figure}[t]
    \centering
    \includegraphics[width=0.98\linewidth]{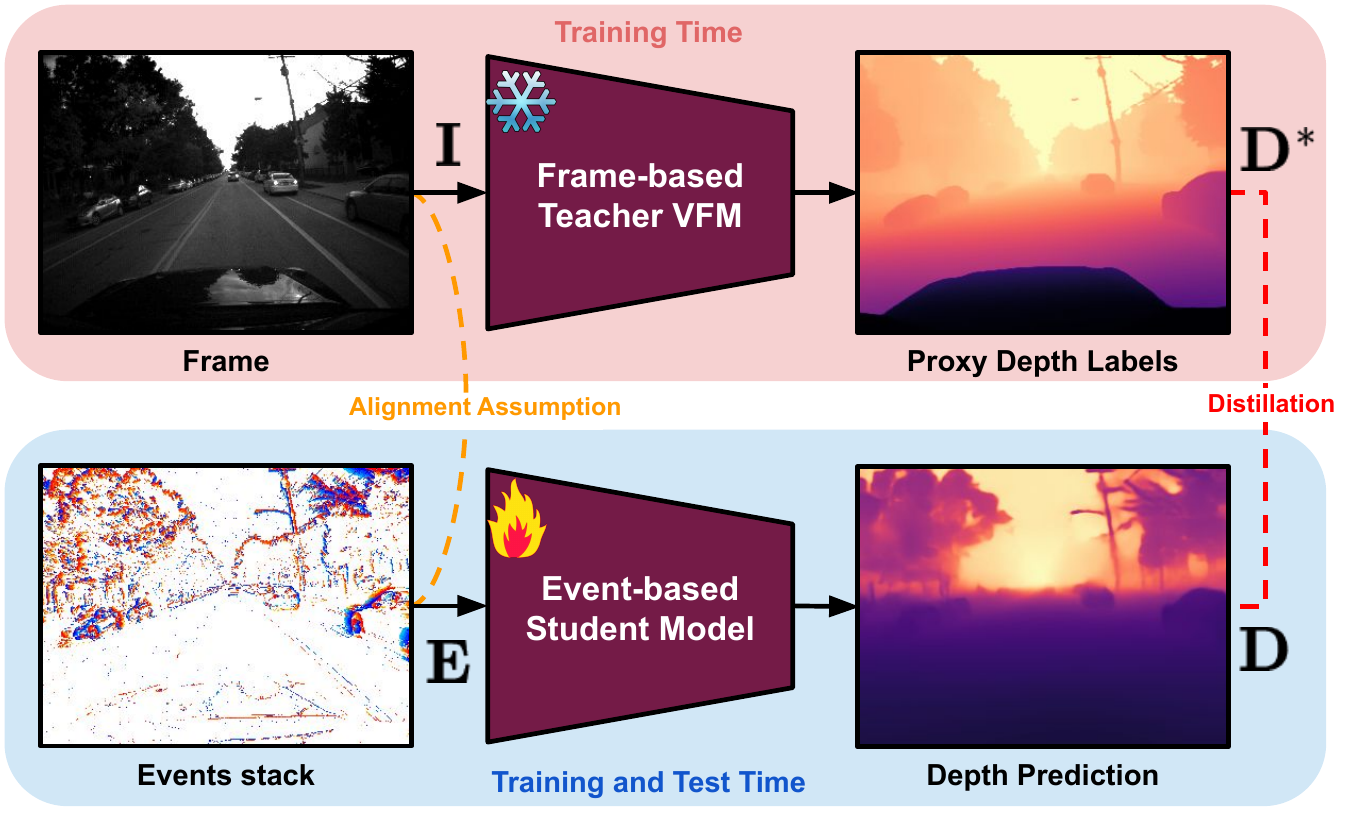}
    \vspace{-0.3cm}\caption{
    \textbf{Proposed Cross-Modal Distillation Strategy.} During training, a VFM teacher processes RGB input frames $\mathbf{I}$ to generate proxy depth labels $\mathbf{D}^*$, which supervise an event-based student model. The student takes aligned event stacks $\mathbf{E}$ as input and predicts the final depth map $\mathbf{D}$.}\vspace{-0.3cm}
    \label{fig:self_framework}
\end{figure}

Our first goal is to leverage the knowledge of frame-based monocular depth models like DAv2 extracting pseudo labels to train \textit{any} event-based student depth model -- \eg, E2Depth -- given aligned intensity frames and event stacks.
Figure \ref{fig:self_framework} outlines our cross-modal distillation paradigm. Moreover, we propose to cast a frame-based model -- DAv2 in our experiments -- either in its original version or enriching it to exploit temporal cues, to the event domain taking advantage of the massive pre-train performed in the image domain.

\begin{figure*}[t]
    \centering
    \includegraphics[clip,trim=0cm 5.2cm 0cm 0cm,width=0.99\linewidth]{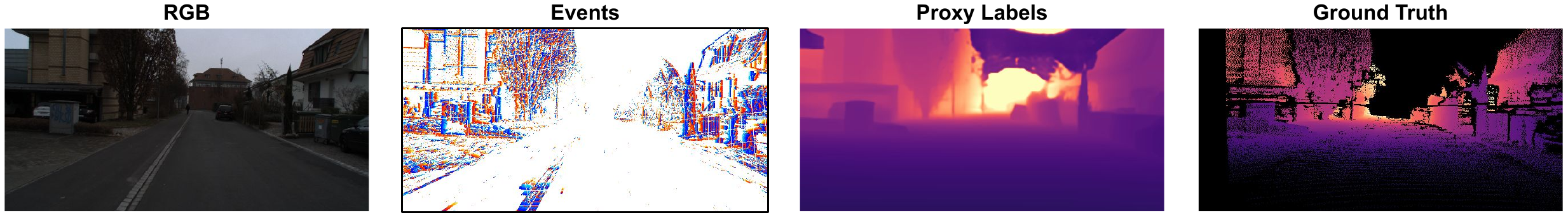}
    \vspace{-0.3cm}\caption{\textbf{Labels Distillation from Frame-Based Vision Foundation Model.} Given the availability of aligned color and event modalities, e.g., collected by a DAVIS346B sensor, we can exploit a VFM to extract proxy labels from the color images, resulting in much dense supervision compared to the one provided by semi-dense LiDAR annotations. }\vspace{-0.3cm}
    \label{fig:proxy_labels}
\end{figure*}

\begin{figure*}[t]
    \centering
    \includegraphics[width=0.95\linewidth]{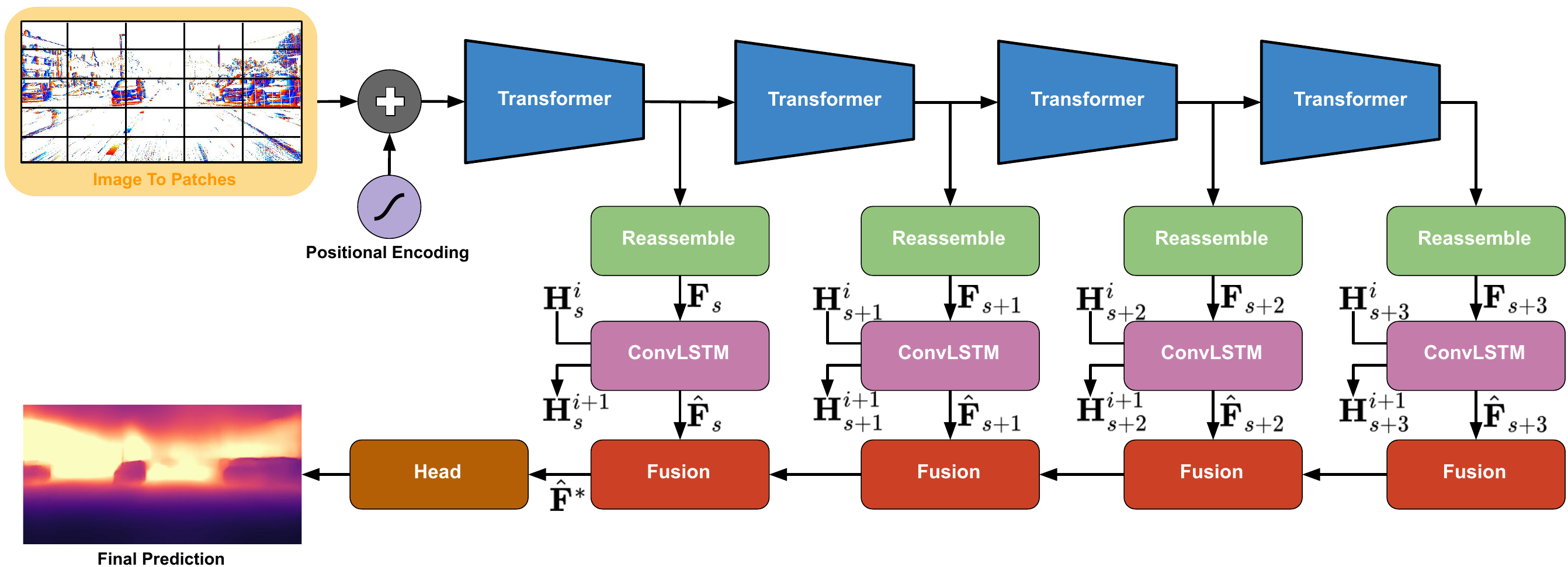} 
    \vspace{-0.3cm}\caption{\textbf{Proposed Recurrent VFM.} Our DepthAnyEvent-R model processes image patches with positional encoding through multiple transformer stages that produce multi-scale feature maps $\mathbf{F}_s$. These features are combined with hidden states $\mathbf{H}_s^i$ in ConvLSTM modules $\mathcal{R}_s$ to incorporate temporal information from previous event stacks, generating enhanced feature maps $\hat{\mathbf{F}}_s$ and updated hidden states $\mathbf{H}_s^{i+1}$. A hierarchical fusion process integrates features from different scales to predict the final depth prediction $\hat{\mathbf{F}}^*$.}\vspace{-0.3cm}
    \label{fig:rec_arch}
\end{figure*}

\subsection{VFMs for Cross-Modal Distillation}

Visual Foundation Models have achieved astonishing results mainly due to their peculiar large-scale training procedures. For instance, DAv2 relies on a DINOv2 backbone that was pre-trained with hundreds of millions of images in an unsupervised manner. Furthermore, DAv2 uses tens of millions of pseudo-labeled and millions of labeled images for training.
Unfortunately, event data lacks equivalent large-scale datasets \cite{mvsec,gehrig2021dsec,chaney2023m3ed}, substantially precluding comparable training in the event domain. To bridge this gap, we propose leveraging a pre-trained VFM -- DAv2 ViT-Large in our experiments-- to provide dense supervision for any event-based depth estimation networks, as outlined in Figure \ref{fig:self_framework}. During training, a teacher VFM processes a frame, producing the proxy label $\mathbf{D}^*$ (Fig. \ref{fig:proxy_labels} shows an example) and the student model predicts a depth map $\mathbf{D}$ from the spatially and temporally aligned events. The student model is supervised using a loss $\mathcal{L}=\mathcal{L}_{si}+\lambda\mathcal{L}_{reg}$ composed of a scale-invariant loss $\mathcal{L}_{si}$ and a gradient regularization term $\mathcal{L}_{reg}$ \cite{MiDas}:
\begin{equation}
    \mathcal{L}_{si}(\hat{\mathbf{D}},\hat{\mathbf{D}}^*) = \frac{1}{2\left|\mathbf{M}\right|}\sum_{(x,y)\in\mathbf{M}}\left(\hat{\mathbf{D}} - \hat{\mathbf{D}}^*\right)^2
    \label{eq:loss_si}
\end{equation}
where $\mathbf{M}$ is the set of valid pixels, $\hat{\mathbf{D}}=s{\mathbf{D}}+t$ and $\hat{\mathbf{D}}^*={\mathbf{D}}^*$ are respectively the scaled and shifted versions of the student prediction $\mathbf{D}$ and the proxy label ${\mathbf{D}}^*$, and $(s,t)$ are the scaling factors obtained using the least-square approach:
\begin{equation}
    (s,t) = \text{arg}\min_{s,t}\sum_{(x,y)\in\mathbf{M}}\left({s\mathbf{D}}+t - {\mathbf{D}}^*\right)^2
    \label{eq:leastsquare}
\end{equation}
The regularization term $\mathcal{L}_{reg}$ is defined as follows:
\begin{equation}
    \mathcal{L}_{reg}(\hat{\mathbf{D}},\hat{\mathbf{D}}^*) = \sum^K_{k=1}\frac{1}{\left|\mathbf{M}_k\right|}\sum_{(x,y)\in\mathbf{M}_k}\left( \left|\nabla_x \mathbf{R}_k\right| + \left|\nabla_y \mathbf{R}_k\right| \right)
    \label{eq:loss_reg}
\end{equation}
where $\mathbf{R}_k=\hat{\mathbf{D}}_k - \hat{\mathbf{D}}^*_k$ is the difference of maps at scale $k$ and $\mathbf{M}_k$ is the set of valid pixels at scale $k$.

To ensure alignment, frame and event cameras must be calibrated -- intrinsically done in the DAVIS camera -- and events are sliced from the frame's acquiring timestamp.

\begin{table*}[t]
    \centering
    \renewcommand{\tabcolsep}{12pt}
    \resizebox{1.0\textwidth}{!} 
    {
        \begin{tabular}{l | c | c  c c c c|c c c }
        \toprule
             \textbf{Model} & \textbf{Dataset} &
             \textbf{Abs Rel}$\downarrow$ &  \textbf{Sq Rel} $\downarrow$ &
             \textbf{RMSE}$\downarrow$ &  \textbf{RMSE log}$\downarrow$ &
             \textbf{SI log}$\downarrow$ & $ \boldsymbol{\delta < 1.25\uparrow}$ &
             $\boldsymbol{\delta < 1.25^2 \uparrow}$  & $\boldsymbol{\delta < 1.25^3 \uparrow}$ \\ 
        \toprule
E2Depth \cite{hidalgo2020learning} &  & {  0.527 } & {  1.122 } & \underline{  7.894 } & {  0.512 } & \underline{  0.244 } & {  0.363 } & {  0.637 } & {  0.811 } \\
EReFormer \cite{liu2024event} & MVSEC & {  0.518 } & {  1.012 } & {  8.423 } & {  0.559 } & {  0.316 } & {  0.361 } & {  0.630 } & {  0.800 } \\
\textbf{DepthAnyEvent} &  & { \bf 0.466 } & \underline{  0.976 } & { \bf 7.824 } & { \bf 0.480 } & { \bf 0.229 } & \underline{  0.408 } & \underline{  0.689 } & { \bf 0.847 } \\
\textbf{DepthAnyEvent-R} &  & \underline{  0.469 } & { \bf 0.946 } & {  8.064 } & \underline{  0.508 } & {  0.272 } & { \bf 0.428 } & { \bf 0.690 } & \underline{  0.832 } \\
\hline
E2Depth \cite{hidalgo2020learning} &  & {  0.395 } & {  0.334 } & {  13.258 } & {  0.412 } & {  0.167 } & {  0.409 } & {  0.719 } & {  0.891 } \\
EReFormer \cite{liu2024event} & DSEC & \underline{  0.297 } & {  0.195 } & {  11.608 } & {  0.334 } & {  0.113 } & \underline{  0.524 } & {  0.824 } & {  0.945 } \\
\textbf{DepthAnyEvent} &  & \underline{  0.297 } & \underline{  0.186 } & \underline{  11.072 } & \underline{  0.330 } & \underline{  0.108 } & {  0.519 } & \underline{  0.827 } & \underline{  0.948 } \\
\textbf{DepthAnyEvent-R} &  & { \bf 0.276 } & { \bf 0.165 } & { \bf 10.942 } & { \bf 0.314 } & { \bf 0.101 } & { \bf 0.555 } & { \bf 0.843 } & { \bf 0.954 } \\
        \bottomrule
        \end{tabular}
    }
    \vspace{-0.3cm}\caption{\textbf{Quantitative Results -- Zero-Shot Generalization on MVSEC and DSEC.} All networks are trained on the EventScape synthetic dataset only, and tested without any fine-tuning. }\vspace{-0.3cm}
    \label{tab:mvsec_synth}
\end{table*}

\begin{figure*}[t]
    \centering
    \includegraphics[width=0.98\linewidth]{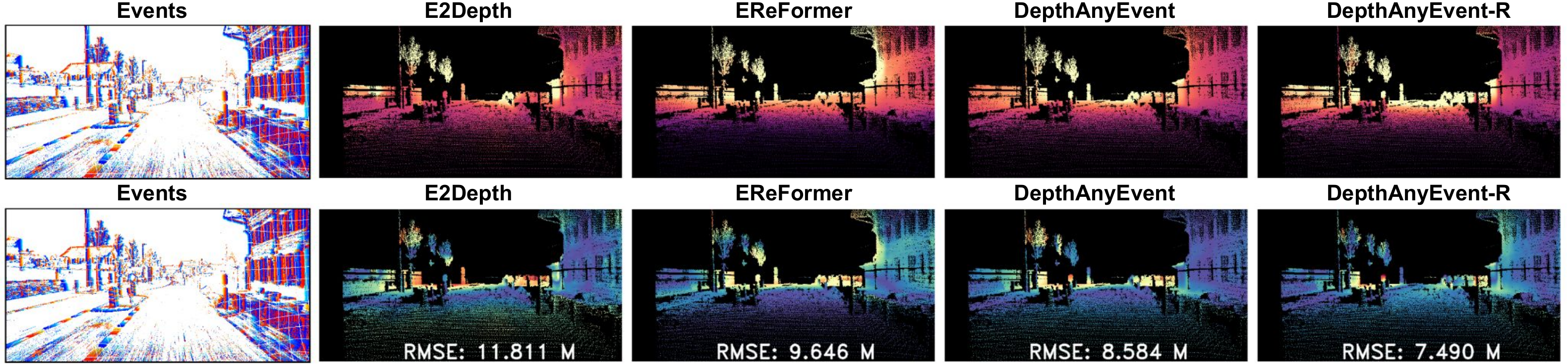}
    \vspace{-0.3cm}\caption{\textbf{Qualitative Results on DSEC dataset -- Zero-Shot Generalization.} From left to right: event image, predictions by E2Depth, EReFormer, DepthAnyEvent and DepthAnyEvent-R, trained on EventScape only.}\vspace{-0.3cm}
    \label{fig:qualitative_synth_mvsec}
\end{figure*}

\subsection{Casting VFMs to the Event Domain}

Frame-based monocular depth models cannot be used directly on events, given the diverse nature of the latter.
Hence, to adapt their capabilities to the event domain, we choose an appropriate event representation that can reduce the gap between frames and events encoding. Furthermore, we exploit the sequential nature of temporal events, proposing a novel recurrent architecture of DAv2.

\textbf{Choosing the Right Event Representation.} The events stream contains spatial and temporal information; hence, a good event representation should capture both to ensure limited loss of information.
Since monocular models naturally process RGB frames -- \ie, they produce a depth map given an image $\mathbf{I} \in \mathbb{R}^{W \times H \times 3}$ as input -- we have to choose an event representation that encodes both spatial and temporal requirements within an RGB frame to pursue minimal modifications of the pre-trained VFM.

Purposely, the Tencode \cite{Huang_2023_WACV} representation fits with our aim. Consequently, starting from a sliced event history $\mathcal{E}_{t_d}$, either using SBT or SBN \cite{nam2022stereo}, Tencode encodes $\mathcal{E}_{t_d}$ into a stack $\mathbf{E}$ as follows:
\begin{equation}
    \mathbf{E}(x_k,y_k) = \begin{cases}
        (1,\frac{t_d-t_k}{\Delta T},0) \ \text{if} \  p_k=1 \\
        (0,\frac{t_d-t_k}{\Delta T},1) \ \text{if} \  p_k=-1 \\
    \end{cases}
    \label{eq:tencode}
\end{equation}
where $e_k= (x_k,y_k,p_k,t_k) \in \mathcal{E}_{t_d}$ is the $k$-th event of $\mathcal{E}_{t_d}$ and $\Delta T$ is the time interval of event slice $\mathcal{E}_{t_d}$.

\textbf{VFM for Events.} Although the Tencode representation significantly differs from a conventional RGB image of the same scene, we propose to adapt a pre-trained VFM to deal with the event domain through fine-tuning with event data using the Tencode representation. For this purpose, we use as the VFM a vanilla DAv2 ViT-S for our experiments. We dubbed the model as \textit{DepthAnyEvent}.

\textbf{Recurrent VFM for Events.} Additionally, given the sequence nature of the event stream, Recurrent Neural Networks (RNNs) could encode previous features extracted from past event stacks into a hidden state \cite{hidalgo2020learning,liu2024event}. At each iteration, the recurrent module can update the hidden state with the features extracted from the current stack, generating a new hidden state for the next iteration.

However, monocular depth models typically lack a recurrent module since they are designed to work with single-frame instances. Hence, for our purposes, this could hinder the quality of predictions, especially during static scenes where events are not triggered. To effectively adapt them to the event domain, we introduce a recurrent extension of DAv2 ViT-Small, dubbed as \textit{DepthAnyEvent-R}, that integrates cues from previous event stacks, as outlined in Figure \ref{fig:rec_arch}. 
The DAv2 architecture is composed of two main modules: a DINOv2 \cite{oquab2024dinov} Encoder $\mathcal{G}$ based on Visual Transformer (ViT), and a Dense Depth Decoder $\mathcal{D}$. Given an image $\mathbf{I}$ encoded with the Tencode representation, the encoder $\mathcal{G}$ first splits the image into patches and adds positional encoding to them. Next, patches are passed through multiple transformer stages and then reassembled from different stages into multi-scale feature maps $\mathbf{F}_s \in \mathbb{R}^{\frac{W}{s} \times \frac{H}{s} \times C_s}$. For each scale $s$, we feed the feature maps $\mathbf{F}_s$ and the hidden state $\mathbf{H}_s^i \in \mathbb{R}^{\frac{W}{s} \times \frac{H}{s} \times C_s}$ with $\mathbf{H}_s^0=\mathbf{0}$ to a ConvLSTM \cite{shi2015convolutional} module $\mathcal{R}_s$ obtaining a new hidden state $\mathbf{H}_s^{i+1}$ and temporally enhanced feature maps ${\hat{\mathbf{F}}_s}$. Starting from the lowest scale, a series of fusion modules sequentially upsample and fuse the feature maps to obtain the final feature map $\hat{\mathbf{F}}^*$ fed to the decoder $\mathcal{D}$ to obtain the final predicted depth map.

\begin{table*}[t!]
    \centering
    \renewcommand{\tabcolsep}{12pt}
    \resizebox{1.0\textwidth}{!} 
    {
        \begin{tabular}{l | c | c  c c c c|c c c }
        \toprule
             \textbf{Model} & \textbf{Dataset} &
             \textbf{Abs Rel}$\downarrow$ &  \textbf{Sq Rel} $\downarrow$ &
             \textbf{RMSE}$\downarrow$ &  \textbf{RMSE log}$\downarrow$ &
             \textbf{SI log}$\downarrow$ & $ \boldsymbol{\delta < 1.25\uparrow}$ &
             $\boldsymbol{\delta < 1.25^2 \uparrow}$  & $\boldsymbol{\delta < 1.25^3 \uparrow}$ \\ 
        \toprule
E2Depth \cite{hidalgo2020learning} &  & {  0.420 } & {  0.806 } & {  7.268 } & \underline{  0.455 } & { \bf 0.213 } & {  0.432 } & {  0.717 } & {  0.868 } \\
EReFormer \cite{liu2024event} & MVSEC & {  0.511 } & {  1.057 } & {  8.373 } & {  0.523 } & {  0.274 } & {  0.391 } & {  0.652 } & {  0.810 } \\
\textbf{DepthAnyEvent} &  & \underline{  0.373 } & \underline{  0.715 } & \underline{  6.627 } & { \bf 0.449 } & \underline{  0.222 } & \underline{  0.471 } & \underline{  0.747 } & { \bf 0.884 } \\
\textbf{DepthAnyEvent-R} &  & { \bf 0.365 } & { \bf 0.691 } & { \bf 6.465 } & {  0.483 } & {  0.258 } & { \bf 0.489 } & { \bf 0.751 } & \underline{  0.878 } \\
\hline
E2Depth \cite{hidalgo2020learning} &  & {  0.253 } & {  0.130 } & {  10.119 } & {  0.315 } & {  0.107 } & {  0.574 } & {  0.861 } & {  0.956 } \\
EReFormer \cite{liu2024event} & DSEC & {  0.286 } & {  0.208 } & {  11.369 } & {  0.325 } & {  0.109 } & {  0.569 } & {  0.839 } & {  0.944 } \\
\textbf{DepthAnyEvent} &  & \underline{  0.201 } & \underline{  0.079 } & \underline{  8.880 } & \underline{  0.266 } & \underline{  0.077 } & \underline{  0.664 } & \underline{  0.917 } & \underline{  0.975 } \\
\textbf{DepthAnyEvent-R} &  & { \bf 0.191 } & { \bf 0.070 } & { \bf 8.618 } & { \bf 0.244 } & { \bf 0.064 } & { \bf 0.691 } & { \bf 0.930 } & { \bf 0.981 } \\
        \bottomrule
        \end{tabular}
    }
    \vspace{-0.2cm}\caption{\textbf{Quantitative Results -- In-Domain Evaluation on MVSEC and DSEC.} All networks are trained on the EventScape synthetic dataset and then further fine-tuned on MVSEC and DSEC datasets separately. }\vspace{-0.2cm}
    \label{tab:finetuning}
\end{table*}

\begin{figure*}[t]
    \centering
    \includegraphics[width=0.98\linewidth]{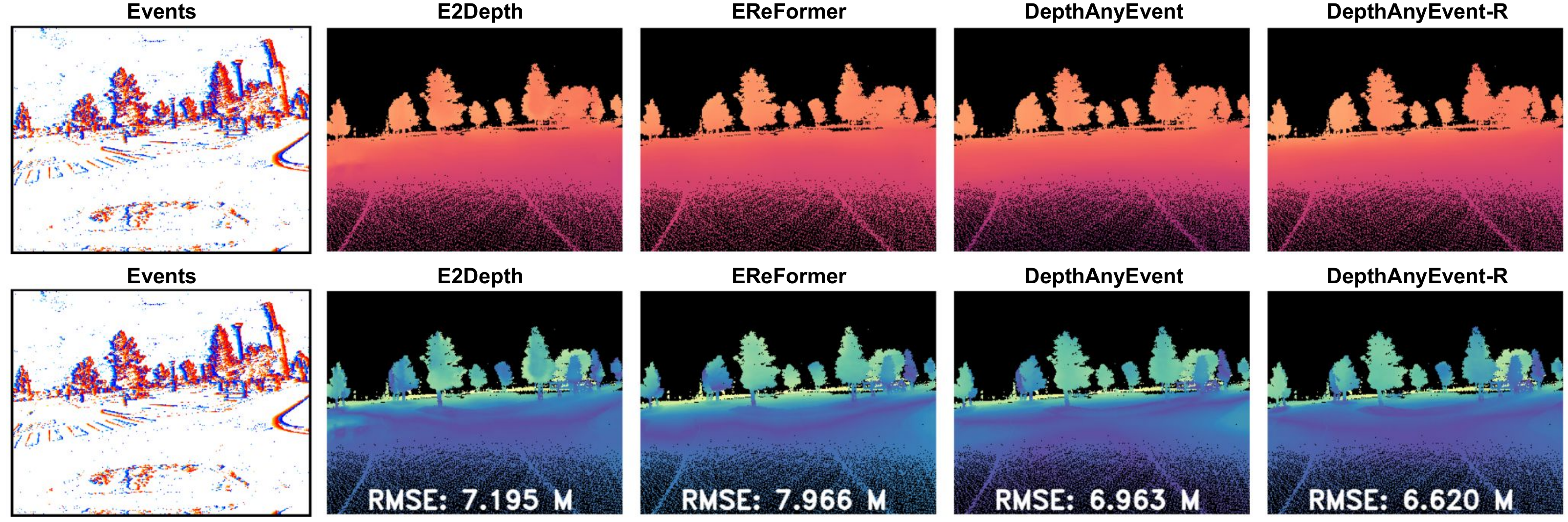}
    \vspace{-0.2cm}\caption{\textbf{Qualitative Results on MVSEC -- Fine-tuned Models.} From left to right: event image, predictions by E2Depth, EReFormer, DepthAnyEvent and DepthAnyEvent-R, trained on EventScape and fine-tuned on MVSEC.}
    \label{fig:qualitative_sup_mvsec}
\end{figure*}

\section{Experiments}
\label{sec:experiments}

We describe our implementation details, datasets, and evaluation protocols, followed by experiments.

\subsection{Implementation and Experimental Settings}

\textbf{Hyperparameters Settings.} We set the slicing window $\Delta T$, the number of Voxel Grid bins $B$, and the loss factor $\lambda$ respectively to 50ms, 5, and 0.25.
We implement event-based student networks E2Depth \cite{hidalgo2020learning} and EReFormer \cite{liu2024event} starting from their codebase. For DepthAnyEvent and DepthAnyEvent-R, we start from the DAv2 ViT-Small codebase \cite{depth_anything_v2}. We use PyTorch, and a single A100 GPU with 64GB of RAM.
Following the original papers, we fix the learning rate to $10^{-4}$ and $3.2\cdot10^{-5}$ respectively for E2Depth and EReFormer, while we set a learning rate of $5\cdot10^{-6}$ for all DepthAnyEvent variants.
We adjust the training steps to 75k, using the AdamW optimizer with the OneCycle scheduler, and apply data augmentations including horizontal flips and random crops at $224 \times 224$. We set the batch size to 10, except for EReFormer: given the higher memory requirements, we change it to 2. We unroll all recurrent networks -- \ie, E2Depth, EReFormer, and DepthAnyEvent-R -- for 20 steps. 
We choose as the event representation Tencode \cite{Huang_2023_WACV} for DepthAnyEvent and DepthAnyEvent-R, while we maintained the original representation for E2Depth and EReFormer -- \ie, respectively, Voxel Grid \cite{zhu2019unsupervised} and Image-like \cite{liu2024event}. 
Finally, we use the scale-invariant $\mathcal{L}$ for all networks.
The settings reported are used for all experiments unless otherwise specified.

\textbf{Proxy Labels Factory.} We generate proxy labels from frames using the DAv2 ViT-Large trained for metric depth estimation: starting from the \textit{Large} vanilla weights provided by the authors, we perform a fine-tuning on EventScape \cite{gehrig2021combining} for 10k steps with a learning rate of $10^{-6}$.

\textbf{Synthetic Training Setup.} We obtain the synthetic checkpoints for all networks training on the synthetic EventScape \cite{gehrig2021combining} dataset. While E2Depth was trained from scratch, we followed EReFormer's original paper and set Swin-T pre-trained on ImageNet as the backbone. For DepthAnyEvent and DepthAnyEvent-R, we started from the \textit{Small} weights provided by the authors.

\textbf{Fine-tuning Setup.} We follow \cite{hidalgo2020learning}, fine-tuning the models to the target domain using both real and synthetic data -- \ie, MVSEC \cite{mvsec} + EventScape \cite{gehrig2021combining}, and DSEC\cite{gehrig2021dsec} + EventScape \cite{gehrig2021combining} -- starting from the synthetic checkpoints obtained in the previous point.

\textbf{Distillation Training Setup.} We use the proxy labels previously generated with DAv2 ViT-L instead of the original sparse ground-truth. Differently from the previous point, we trained the models on the dense proxy labels only instead of a synthetic+proxy mixture.

\begin{table*}[t!]
    \centering
    \renewcommand{\tabcolsep}{12pt}
    \resizebox{1.0\textwidth}{!} 
    {
        \begin{tabular}{l | c | c  c c c c|c c c }
        \toprule
             \textbf{Model} & \textbf{Dataset} &
             \textbf{Abs Rel}$\downarrow$ &  \textbf{Sq Rel} $\downarrow$ &
             \textbf{RMSE}$\downarrow$ &  \textbf{RMSE log}$\downarrow$ &
             \textbf{SI log}$\downarrow$ & $ \boldsymbol{\delta < 1.25\uparrow}$ &
             $\boldsymbol{\delta < 1.25^2 \uparrow}$  & $\boldsymbol{\delta < 1.25^3 \uparrow}$ \\ 
        \toprule
E2Depth Synth & MVSEC & {  0.527 } & {  1.122 } & {  7.894 } & \underline{  0.512 } & \underline{  0.244 } & {  0.363 } & {  0.637 } & {  0.811 } \\
\textbf{E2Depth Distilled} &  & { \bf 0.400 } & \underline{  0.817 } & { \bf 6.786 } & {  0.538 } & {  0.304 } & { \bf 0.479 } & { \bf 0.740 } & \underline{  0.865 } \\
E2Depth Supervised &  & \underline{  0.420 } & { \bf 0.806 } & \underline{  7.268 } & { \bf 0.455 } & { \bf 0.213 } & \underline{  0.432 } & \underline{  0.717 } & { \bf 0.868 } \\
\hline
EReFormer Synth & MVSEC & {  0.518 } & \underline{  1.012 } & {  8.423 } & {  0.559 } & {  0.316 } & {  0.361 } & {  0.630 } & {  0.800 } \\
\textbf{EReFormer Distilled} &  & { \bf 0.448 } & { \bf 0.817 } & { \bf 7.867 } & { \bf 0.498 } & { \bf 0.253 } & { \bf 0.434 } & { \bf 0.700 } & { \bf 0.842 } \\
EReFormer Supervised &  & \underline{  0.511 } & {  1.057 } & \underline{  8.373 } & \underline{  0.523 } & \underline{  0.274 } & \underline{  0.391 } & \underline{  0.652 } & \underline{  0.810 } \\
\hline
DepthAnyEvent Synth & MVSEC & {  0.466 } & {  0.976 } & {  7.824 } & \underline{  0.480 } & \underline{  0.229 } & {  0.408 } & {  0.689 } & {  0.847 } \\
\textbf{DepthAnyEvent Distilled} &  & \underline{  0.397 } & \underline{  0.771 } & \underline{  6.910 } & {  0.495 } & {  0.260 } & \underline{  0.461 } & \underline{  0.735 } & \underline{  0.870 } \\
DepthAnyEvent Supervised &  & { \bf 0.373 } & { \bf 0.715 } & { \bf 6.627 } & { \bf 0.449 } & { \bf 0.222 } & { \bf 0.471 } & { \bf 0.747 } & { \bf 0.884 } \\
\hline
DepthAnyEvent-R Synth & MVSEC & {  0.469 } & {  0.946 } & {  8.064 } & \underline{  0.508 } & \underline{  0.272 } & {  0.428 } & {  0.690 } & {  0.832 } \\
\textbf{DepthAnyEvent-R Distilled} &  & \underline{  0.399 } & \underline{  0.781 } & \underline{  6.830 } & {  0.509 } & {  0.281 } & \underline{  0.462 } & \underline{  0.735 } & \underline{  0.866 } \\
DepthAnyEvent-R Supervised &  & { \bf 0.365 } & { \bf 0.691 } & { \bf 6.465 } & { \bf 0.483 } & { \bf 0.258 } & { \bf 0.489 } & { \bf 0.751 } & { \bf 0.878 } \\
\hline\hline
E2Depth Synth & DSEC & {  0.395 } & {  0.334 } & {  13.258 } & {  0.412 } & {  0.167 } & {  0.409 } & {  0.719 } & {  0.891 } \\
\textbf{E2Depth Distilled} &  & \underline{  0.272 } & \underline{  0.153 } & \underline{  10.579 } & { \bf 0.309 } & { \bf 0.096 } & \underline{  0.551 } & \underline{  0.851 } & { \bf 0.959 } \\
E2Depth Supervised &  & { \bf 0.253 } & { \bf 0.130 } & { \bf 10.119 } & \underline{  0.315 } & \underline{  0.107 } & { \bf 0.574 } & { \bf 0.861 } & \underline{  0.956 } \\
\hline
EReFormer Synth & DSEC & {  0.297 } & { \bf 0.195 } & {  11.608 } & {  0.334 } & {  0.113 } & {  0.524 } & {  0.824 } & { \bf 0.945 } \\
\textbf{EReFormer Distilled} &  & { \bf 0.285 } & \underline{  0.198 } & \underline{  11.407 } & \underline{  0.327 } & \underline{  0.111 } & \underline{  0.563 } & { \bf 0.839 } & \underline{  0.944 } \\
EReFormer Supervised &  & \underline{  0.286 } & {  0.208 } & { \bf 11.369 } & { \bf 0.325 } & { \bf 0.109 } & { \bf 0.569 } & \underline{  0.839 } & {  0.944 } \\
\hline
DepthAnyEvent Synth & DSEC & {  0.297 } & {  0.186 } & {  11.072 } & {  0.330 } & {  0.108 } & {  0.519 } & {  0.827 } & {  0.948 } \\
\textbf{DepthAnyEvent Distilled} &  & \underline{  0.213 } & \underline{  0.095 } & \underline{  8.930 } & { \bf 0.253 } & { \bf 0.065 } & \underline{  0.662 } & \underline{  0.915 } & { \bf 0.980 } \\
DepthAnyEvent Supervised &  & { \bf 0.201 } & { \bf 0.079 } & { \bf 8.880 } & \underline{  0.266 } & \underline{  0.077 } & { \bf 0.664 } & { \bf 0.917 } & \underline{  0.975 } \\
\hline
DepthAnyEvent-R Synth & DSEC & {  0.276 } & {  0.165 } & {  10.942 } & {  0.314 } & {  0.101 } & {  0.555 } & {  0.843 } & {  0.954 } \\
\textbf{DepthAnyEvent-R Distilled} &  & \underline{  0.226 } & \underline{  0.111 } & \underline{  9.310 } & \underline{  0.266 } & \underline{  0.072 } & \underline{  0.638 } & \underline{  0.906 } & \underline{  0.977 } \\
DepthAnyEvent-R Supervised &  & { \bf 0.191 } & { \bf 0.070 } & { \bf 8.618 } & { \bf 0.244 } & { \bf 0.064 } & { \bf 0.691 } & { \bf 0.930 } & { \bf 0.981 } \\
        \bottomrule
        \end{tabular}
    }
    \vspace{-0.2cm}\caption{\textbf{Quantitative Results -- Supervised vs Distilled Models on MVSEC and DSEC.} All networks are trained on the EventScape synthetic dataset and then fine-tuned on MVSEC and DSEC datasets separately, either through distillation or on ground-truth depth labels.}\vspace{-0.2cm}
    \label{tab:mvsec_all}
\end{table*}

\begin{figure*}[t]
    \centering
    \includegraphics[width=0.98\linewidth]{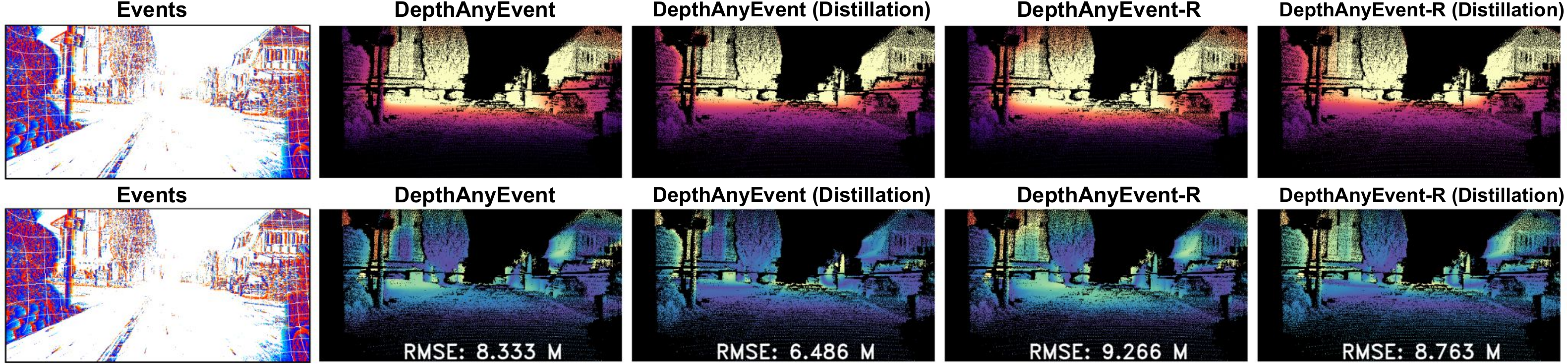}
    \vspace{-0.2cm}\caption{\textbf{Qualitative Results on DSEC -- Supervised vs Distilled Models.} From left to right: event image, predictions by DepthAnyEvent and its distilled counterpart, and by DepthAnyEvent-R and its distilled counterpart.}\vspace{-0.2cm}
    \label{fig:qualitative_self_dsec}
\end{figure*}

\begin{table*}[t]
    \centering
    \renewcommand{\tabcolsep}{20pt}
    \resizebox{1.0\textwidth}{!} 
    {
        \begin{tabular}{l | c  c c c c|c c c }
        \toprule
             \textbf{Model} & 
             \textbf{Abs Rel}$\downarrow$ &  \textbf{Sq Rel} $\downarrow$ &
             \textbf{RMSE}$\downarrow$ &  \textbf{RMSE log}$\downarrow$ &
             \textbf{SI log}$\downarrow$ & $ \boldsymbol{\delta < 1.25\uparrow}$ &
             $\boldsymbol{\delta < 1.25^2 \uparrow}$  & $\boldsymbol{\delta < 1.25^3 \uparrow}$ \\ 
        \toprule
        E2Depth [{\color{iccvblue} 15}] &  {  0.344 } & {  0.253 } & {  13.467 } & {  0.376 } & {  0.098 } & {  0.447 } & {  0.755 } & {  0.915 } \\
        EReFormer [{\color{iccvblue} 21}] & {  0.387 } & {  0.401 } & {  13.954 } & {  0.395 } & {  0.124 } & {  0.486 } & {  0.776 } & {  0.892 } \\
        \textbf{DepthAnyEvent} & \underline{  0.277 } & \underline{  0.170 } & \underline{  11.117 } & \underline{  0.292 } & \underline{  0.051 } & \underline{  0.585 } & \underline{  0.860 } & \underline{  0.955 } \\
        \textbf{DepthAnyEvent-R} &  { \bf 0.252 } & { \bf 0.128 } & { \bf 9.824 } & { \bf 0.268 } & { \bf 0.045 } & { \bf 0.592 } & { \bf 0.900 } & { \bf 0.971 } \\
        \bottomrule
        \end{tabular}
    }
    \vspace{-0.3cm}
    \caption{\textbf{Metric Depth Evaluation.} Training and evaluation on DSEC dataset.}\vspace{-0.2cm}
    \label{tab:metric_depth}
\end{table*}

\begin{table*}[t]
    \centering
    \renewcommand{\tabcolsep}{5pt}
    \resizebox{1.0\textwidth}{!} 
    {
        \begin{tabular}{c l l l | c  c c c c|c c c }
        \toprule
             & \textbf{Model} & \textbf{Supervision} & \textbf{Experiment} &
             \textbf{Abs Rel}$\downarrow$ &  \textbf{Sq Rel} $\downarrow$ &
             \textbf{RMSE}$\downarrow$ &  \textbf{RMSE log}$\downarrow$ &
             \textbf{SI log}$\downarrow$ & $ \boldsymbol{\delta < 1.25\uparrow}$ &
             $\boldsymbol{\delta < 1.25^2 \uparrow}$  & $\boldsymbol{\delta < 1.25^3 \uparrow}$ \\ 
        \toprule
        (A) & \multirow{2}{*}{DepthAnyEvent-R} & Distillation & Tencode+DAv2  & {\bf  0.399 } & {\bf   0.781 } & {\bf   6.830 } & {  0.509 } & {  0.281 } & {\bf   0.462 } & {\bf   0.735 } & {  0.866 } \\
        (B) & & Distillation & Tencode+DepthPro  & {  0.429 } & {  0.942 } & {  7.472 } & {\bf   0.452 } & {\bf   0.208 } & {  0.444 } & {  0.726 } & {\bf   0.869 } \\
        \hline
        (C) & \multirow{4}{*}{DepthAnyEvent-R } & Ground-truth & Tencode+DAv2 & {  0.365 } & {\bf   0.691 } & {\bf   6.465 } & {  0.483 } & {  0.258 } & {  0.489 } & {  0.751 } & {  0.878 } \\
        (D) & & Ground-truth & VoxelGrid+DAv2 & {  0.382 } & {  0.719 } & {  6.932 } & {  0.444 } & {  0.215 } & {  0.473 } & {  0.742 } & {  0.877 } \\
        (E) & & Ground-truth & Tencode+DAv2 (no pretrain) & {  0.446 } & {  0.799 } & {  7.492 } & {  0.506 } & {  0.260 } & {  0.390 } & {  0.678 } & {  0.845 } \\
        (F) & & {Ground-truth + Distillation} & Tencode+DAv2 &  {\bf   0.362 } & {  0.697 } & {  6.511 } & {\bf   0.438 } & {\bf   0.211 } & {\bf   0.494 } & {\bf   0.760 } & {\bf   0.890 } \\
        \bottomrule 
        \end{tabular}
    }
    \vspace{-0.3cm}
    \caption{\textbf{Ablation Studies.} Training and evaluation on MVSEC dataset.}\vspace{-0.3cm}
    \label{tab:additional_ablation}
\end{table*}

\begin{table}[t]
    \centering
    \renewcommand{\tabcolsep}{20pt}
    \resizebox{0.48\textwidth}{!}{ 
    \begin{tabular}{l|cc}
        \toprule
        Model & Inference (ms) & Memory (MB)  \\
        \hline
        E2Depth [{\color{iccvblue} 15}] & 1.50 & 242 \\
        EReFormer [{\color{iccvblue} 21}] & 35.75 & 534 \\
        {\bf DepthAnyEvent} & \bf 1.26 & \bf 71 \\
        {\bf DepthAnyEvent-R} & 9.20 & 202 \\
        \bottomrule
    \end{tabular}}
    \vspace{-0.3cm}
    \caption{\textbf{Computational Analysis.} Inference time on A100 GPU.}\vspace{-0.3cm}
    \label{tab:hw_analysis}
\end{table}

\subsection{Evaluation Datasets \& Protocol}

\textbf{Datasets.} We utilize EventScape \cite{gehrig2021combining} as the synthetic training set, comprising about 120k groundtruth depth maps at resolution of $512\times256$, captured from CARLA \cite{dosovitskiy2017carla} simulator.
For evaluation and domain fine-tunings we used two main benchmarks: MVSEC \cite{mvsec} and DSEC \cite{gehrig2021dsec}. 
The dataset provides events at a resolution of $346\times260$ pixels from a stereo event camera consisting of two DAVIS346B sensors, which also capture spatially aligned images. ground-truth is obtained by processing data from a 16-line LiDAR using Lidar Odometry and Mapping (LOAM), yielding a total of 10k training samples and 20k testing samples. The test set is divided into a 5k-sample daytime subset and three nighttime subsets, each containing 5k samples.
DSEC \cite{gehrig2021dsec} employs two $640\times480$ Prophesee Gen3.1 event cameras in a stereo configuration. Ground-truth disparity is obtained using a 32-line LiDAR, processed with a Lidar Inertial Odometry algorithm, and further filtered to remove outliers.
We convert the disparity ground-truth to depth based on the stereo setup parameters. Unlike MVSEC, RGB frames are captured using a pair of FLIR Blackfly S cameras. To align frames and events, we warp the RGB frames using the calibration parameters. We also apply a $640\times320$ center crop to mitigate misalignment artifacts in nearby objects.
The dataset counts 26k training samples, divided as in \cite{bartolomei2024lidar} into 19k for training and 7k for testing.

\textbf{Evaluation Metrics.} We evaluate the networks using different metrics: absolute relative error (Abs Rel), square Abs Rel (Sq Rel), root mean squared error (RMSE), logarithmic RMSE (RMSE log), logarithmic scale invariant error (SI log), and accuracy with different thresholds ($\delta<1.25$, $\delta<1.25^2$, and $\delta<1.25^3$).
We apply scale and shift to align predictions with the ground-truth before computing the metrics.
We highlight using \textbf{bold} and \underline{underline} the best and second best scores.

\subsection{Synthetic-to-Real Generalization}

We start by evaluating the capability of the different depth estimation models to generalize from synthetic data to real event streams. Purposely, we train E2Depth, EReFormer, DepthAnyEvent, and DepthAnyEvent-R on EventScape and measure their accuracy on both MVSEC and DSEC datasets. Table \ref{tab:mvsec_synth} collects the outcome of this experiment. DepthAnyEvent and DepthAnyEvent-R achieve the best results on almost any metric, hinting how the web-scale training infused in the weights we used to initialize these models represents a solid prior for depth estimation, although coming from images, i.e., a completely different modality with respect to event streams. The two models achieve mixed results one against the other on MVSEC, while DepthAnyEvent-R consistently achieves the best generalization results over DSEC, giving a first intuition about the effectiveness of our design choice to deal with streamed event data. 
Figure \ref{fig:qualitative_synth_mvsec} presents a qualitative comparison of predictions from different models, showcasing the superior zero-shot capabilities of our DepthAnyEvent and DepthAnyEvent-R models.

\subsection{Supervised Fine-tuning}

We now evaluate the accuracy of each model when trained on real event data annotated with semi-dense ground-truth depth. To this aim, we take the weights obtained after training on EventScape and perform further fine-tuning on MVSEC and DSEC separately, then evaluating on the corresponding validation sets. Table \ref{tab:finetuning} reports the results of this evaluation. We can notice, once again, the notable gap in performance between DepthAnyEvent and DepthAnyEvent-R against existing methods EReFormer and E2Depth, confirming again the strong advantage that our models can exploit from the cross-modal training being conducted for image-based depth estimation. Specifically, this time we can notice how DepthAnyEvent-R consistently outperforms the vanilla DepthAnyEvent model on both MVSEC and DSEC datasets, validating our proposed design tailored to event-based depth estimation.

Figure \ref{fig:qualitative_sup_mvsec} shows a qualitative comparison between the predictions by the different models, highlighting the superior accuracy achieved by DepthAnyEvent and, even higher, by DepthAnyEvent-R.

\subsection{Cross-Modal Distillation}

We now assess the effectiveness of our cross-modal distillation strategy compared to conventional, supervised training requiring the availability of costly depth annotations from active sensors. 
Table \ref{tab:mvsec_all} collects the results achieved by each model under the training configuration considered so far, as well as after being trained according to our distillation approach. In most cases, we can notice how the models trained through distillation are comparable, and sometimes even better than their supervised counterparts. 

Figure \ref{fig:qualitative_self_dsec} show some qualitative examples from the DSEC dataset, comparing the predictions by DepthAnyEvent and DepthAnyEvent-R when trained with ground-truth or through distillation. In both cases, distilled models are even more accurate than those supervised with ground-truth.

\subsection{Metric Depth Evaluation}

Finally, we assess the accuracy of our models when trained to predict metric rather than affine-invariant depth. 
Table \ref{tab:metric_depth} collects the results achieved by existing networks and ours when trained on the DSEC dataset for metric depth prediction, evaluated on the validation set of the very same dataset. 
We can appreciate how our two architectures achieve the best results, with DepthAnyEvent-R consistently yielding the best results on any evaluation metrics.

\subsection{Ablation Studies}

We conclude with a study about the impact of different modules in our framework. In the former case, we train different instances of DepthAnyEvent-R on the MVSEC dataset and evaluate on its validation set. Results are collected in Table \ref{tab:additional_ablation}, with row (A) representing the configuration used in the previous experiments.

\textbf{Different VFMs for distillation.} Row (B) shows that replacing Depth Anything v2 with a different VFM for distillation -- i.e., Depth Pro -- yields close results, although slightly worse on most metrics.

\textbf{Input representation.} In rows (C) and (D), we report the results achieved by training our model with ground-truth labels, when processing either Tencode or a voxel-grid representation used to encode raw events. The former yields almost consistently better results.

\textbf{Pre-training.} By training our model starting from DAv2 pretrained weights, we can greatly improve its performance. Indeed, when training DepthAnyEvent-R from scratch (E), the accuracy consistently drops.

\textbf{Combining distillation with ground-truth labels.} Finally, we show how deploying both our cross-modal distillation paradigm and ground-truth annotations (when available) further improves the final model on most metrics.

\subsection{Runtime and Memory Requirements} 

Table \ref{tab:hw_analysis} reports a computational analysis for any model involved in our evaluation. DepthAnyEvent achieves the fastest predictions, using as few as 80MB for a single inference. E2Depth exposes a very similar inference time, although requiring nearly 4$\times$ the memory, while EReFormer runs consistently slower and increases the memory usage to up to 0.5GB. Compared to DepthAnyEvent, the DepthAnyEvent-R variant runs slower, yet still in real-time, and yields more accurate predictions.

\section{Conclusions}
\label{sec:conclusions}
In this paper, we presented a novel approach to event-based monocular depth estimation that leverages the power of pre-trained Visual Foundation Models. Our cross-modal distillation strategy effectively transfers knowledge from frame-based models to the event domain, addressing the crucial challenge of limited ground truth data for event cameras. Experimental results with synthetic and real-world datasets validate our method, showing competitive performance compared to fully supervised methods without requiring expensive depth annotations. Moreover, we have demonstrated two effective methods for adapting VFMs to event data: a vanilla adaptation and a recurrent architecture that better captures the nature of event streams, yielding state-of-the-art performance.

\small{\textbf{Acknowledgment.} This study was carried out within the MOST – Sustainable Mobility National Research Center and received funding from the European Union Next-GenerationEU – PIANO NAZIONALE DI RIPRESA E RESILIENZA (PNRR) – MISSIONE 4 COMPONENTE 2, INVESTIMENTO 1.4 – D.D. 1033 17/06/2022, CN00000023. This manuscript reflects only the authors’ views and opinions, neither the European Union nor the European Commission can be considered responsible for them.

We also acknowledge the CINECA award under the ISCRA initiative for the availability of high-performance computing resources and support.}

{
    \small
    \bibliographystyle{ieeenat_fullname}
    \bibliography{main,old_bib}

\begin{thebibliography}{41}
\providecommand{\natexlab}[1]{#1}
\providecommand{\url}[1]{\texttt{#1}}
\expandafter\ifx\csname urlstyle\endcsname\relax
  \providecommand{\doi}[1]{doi: #1}\else
  \providecommand{\doi}{doi: \begingroup \urlstyle{rm}\Url}\fi

\bibitem[Bartolomei et~al.(2024)Bartolomei, Poggi, Conti, and Mattoccia]{bartolomei2024lidar}
Luca Bartolomei, Matteo Poggi, Andrea Conti, and Stefano Mattoccia.
\newblock Lidar-event stereo fusion with hallucinations.
\newblock In \emph{European Conference on Computer Vision}, pages 125--145. Springer, 2024.

\bibitem[Chaney et~al.(2023)Chaney, Cladera, Wang, Bisulco, Hsieh, Korpela, Kumar, Taylor, and Daniilidis]{chaney2023m3ed}
Kenneth Chaney, Fernando Cladera, Ziyun Wang, Anthony Bisulco, M~Ani Hsieh, Christopher Korpela, Vijay Kumar, Camillo~J Taylor, and Kostas Daniilidis.
\newblock M3ed: Multi-robot, multi-sensor, multi-environment event dataset.
\newblock In \emph{2023 IEEE/CVF Conference on Computer Vision and Pattern Recognition Workshops (CVPRW)}, pages 4016--4023. IEEE, 2023.

\bibitem[Cui et~al.(2022)Cui, Zhu, Liu, Liu, Chen, and Huang]{9686362}
Mingyue Cui, Yuzhang Zhu, Yechang Liu, Yunchao Liu, Gang Chen, and Kai Huang.
\newblock Dense depth-map estimation based on fusion of event camera and sparse lidar.
\newblock \emph{IEEE Transactions on Instrumentation and Measurement}, 71:\penalty0 1--11, 2022.

\bibitem[Dosovitskiy et~al.(2017)Dosovitskiy, Ros, Codevilla, Lopez, and Koltun]{dosovitskiy2017carla}
Alexey Dosovitskiy, German Ros, Felipe Codevilla, Antonio Lopez, and Vladlen Koltun.
\newblock Carla: An open urban driving simulator.
\newblock In \emph{Conference on robot learning}, pages 1--16. PMLR, 2017.

\bibitem[Duan et~al.(2023)Duan, Guo, and Zhu]{Duan2023}
Yiqun Duan, Xianda Guo, and Zheng Zhu.
\newblock {DiffusionDepth}: Diffusion denoising approach for monocular depth estimation.
\newblock \emph{arXiv preprint arXiv:2303.05021}, 2023.

\bibitem[Eigen et~al.(2014)Eigen, Puhrsch, and Fergus]{eigen2014depth}
David Eigen, Christian Puhrsch, and Rob Fergus.
\newblock Depth map prediction from a single image using a multi-scale deep network.
\newblock In \emph{Advances in Neural Information Processing Systems}. Curran Associates, Inc., 2014.

\bibitem[Gallego et~al.(2022)Gallego, Delbruck, Orchard, Bartolozzi, Taba, Censi, Leutenegger, Davison, Conradt, Daniilidis, and Scaramuzza]{EVENT_SURVEY}
Guillermo Gallego, Tobi Delbruck, Garrick~Michael Orchard, Chiara Bartolozzi, Brian Taba, Andrea Censi, Stefan Leutenegger, Andrew Davison, Jorg Conradt, Kostas Daniilidis, and Davide Scaramuzza.
\newblock Event-based vision: A survey.
\newblock \emph{IEEE Transactions on Pattern Analysis and Machine Intelligence}, pages 154--180, 2022.

\bibitem[Gehrig et~al.(2021{\natexlab{a}})Gehrig, R{\"u}egg, Gehrig, Hidalgo-Carri{\'o}, and Scaramuzza]{gehrig2021combining}
Daniel Gehrig, Michelle R{\"u}egg, Mathias Gehrig, Javier Hidalgo-Carri{\'o}, and Davide Scaramuzza.
\newblock Combining events and frames using recurrent asynchronous multimodal networks for monocular depth prediction.
\newblock \emph{IEEE Robotics and Automation Letters}, 6\penalty0 (2):\penalty0 2822--2829, 2021{\natexlab{a}}.

\bibitem[Gehrig et~al.(2021{\natexlab{b}})Gehrig, Aarents, Gehrig, and Scaramuzza]{gehrig2021dsec}
Mathias Gehrig, Willem Aarents, Daniel Gehrig, and Davide Scaramuzza.
\newblock Dsec: A stereo event camera dataset for driving scenarios.
\newblock \emph{IEEE Robotics and Automation Letters}, 6\penalty0 (3):\penalty0 4947--4954, 2021{\natexlab{b}}.

\bibitem[Geiger et~al.(2012)Geiger, Lenz, and Urtasun]{Geiger2012CVPR}
Andreas Geiger, Philip Lenz, and Raquel Urtasun.
\newblock Are we ready for autonomous driving? the {KITTI} vision benchmark suite.
\newblock In \emph{Conference on Computer Vision and Pattern Recognition (CVPR)}, 2012.

\bibitem[Ghosh and Gallego(2024)]{ghosh2024event}
Suman Ghosh and Guillermo Gallego.
\newblock Event-based stereo depth estimation: A survey.
\newblock \emph{arXiv preprint arXiv:2409.17680}, 2024.

\bibitem[Godard et~al.(2016)Godard, Aodha, and Brostow]{godard2017unsupervised}
Cl{\'{e}}ment Godard, Oisin~Mac Aodha, and Gabriel~J. Brostow.
\newblock Unsupervised monocular depth estimation with left-right consistency.
\newblock \emph{CoRR}, abs/1609.03677, 2016.

\bibitem[Godard et~al.(2018)Godard, Aodha, and Brostow]{DBLP:journals/corr/abs-1806-01260}
Cl{\'{e}}ment Godard, Oisin~Mac Aodha, and Gabriel~J. Brostow.
\newblock Digging into self-supervised monocular depth estimation.
\newblock \emph{CoRR}, abs/1806.01260, 2018.

\bibitem[Guizilini et~al.(2023)Guizilini, Vasiljevic, Chen, Ambruș, and Gaidon]{Guizilini2023}
Vitor Guizilini, Igor Vasiljevic, Dian Chen, Rareș Ambruș, and Adrien Gaidon.
\newblock Towards zero-shot scale-aware monocular depth estimation.
\newblock In \emph{ICCV}, 2023.

\bibitem[Hidalgo{-}Carri{\'{o}} et~al.(2020)Hidalgo{-}Carri{\'{o}}, Gehrig, and Scaramuzza]{hidalgo2020learning}
Javier Hidalgo{-}Carri{\'{o}}, Daniel Gehrig, and Davide Scaramuzza.
\newblock Learning monocular dense depth from events.
\newblock \emph{CoRR}, abs/2010.08350, 2020.

\bibitem[Huang et~al.(2023)Huang, Sun, Zhao, Li, and Su]{Huang_2023_WACV}
Ze Huang, Li Sun, Cheng Zhao, Song Li, and Songzhi Su.
\newblock Eventpoint: Self-supervised interest point detection and description for event-based camera.
\newblock In \emph{Proceedings of the IEEE/CVF Winter Conference on Applications of Computer Vision (WACV)}, pages 5396--5405, 2023.

\bibitem[Ji et~al.(2023)Ji, Chen, Xie, Hong, Liu, Liu, Lu, Li, and Luo]{Ji2023}
Yuanfeng Ji, Zhe Chen, Enze Xie, Lanqing Hong, Xihui Liu, Zhaoqiang Liu, Tong Lu, Zhenguo Li, and Ping Luo.
\newblock {DDP}: Diffusion model for dense visual prediction.
\newblock In \emph{ICCV}, 2023.

\bibitem[Laina et~al.(2016)Laina, Rupprecht, Belagiannis, Tombari, and Navab]{laina2016deeper}
Iro Laina, Christian Rupprecht, Vasileios Belagiannis, Federico Tombari, and Nassir Navab.
\newblock Deeper depth prediction with fully convolutional residual networks.
\newblock In \emph{2016 Fourth international conference on 3D vision (3DV)}, pages 239--248. IEEE, 2016.

\bibitem[Lasinger et~al.(2019)Lasinger, Ranftl, Schindler, and Koltun]{MiDas}
Katrin Lasinger, Ren{\'{e}} Ranftl, Konrad Schindler, and Vladlen Koltun.
\newblock Towards robust monocular depth estimation: Mixing datasets for zero-shot cross-dataset transfer.
\newblock \emph{CoRR}, abs/1907.01341, 2019.

\bibitem[Li and Snavely(2018)]{li2018megadepth}
Zhengqi Li and Noah Snavely.
\newblock Megadepth: Learning single-view depth prediction from internet photos.
\newblock In \emph{Proceedings of the IEEE conference on computer vision and pattern recognition}, pages 2041--2050, 2018.

\bibitem[Liu et~al.(2024)Liu, Li, Shi, Fan, Tian, and Zhao]{liu2024event}
Xu Liu, Jianing Li, Jinqiao Shi, Xiaopeng Fan, Yonghong Tian, and Debin Zhao.
\newblock Event-based monocular depth estimation with recurrent transformers.
\newblock \emph{IEEE Transactions on Circuits and Systems for Video Technology}, 34\penalty0 (8):\penalty0 7417--7429, 2024.

\bibitem[Nam et~al.(2022)Nam, Mostafavi, Yoon, and Choi]{nam2022stereo}
Yeongwoo Nam, Mohammad Mostafavi, Kuk-Jin Yoon, and Jonghyun Choi.
\newblock Stereo depth from events cameras: Concentrate and focus on the future.
\newblock In \emph{Proceedings of the IEEE/CVF conference on computer vision and pattern recognition}, pages 6114--6123, 2022.

\bibitem[Nathan~Silberman and Fergus(2012)]{Silberman2012}
Pushmeet~Kohli Nathan~Silberman, Derek~Hoiem and Rob Fergus.
\newblock Indoor segmentation and support inference from rgbd images.
\newblock In \emph{ECCV}, 2012.

\bibitem[Oquab et~al.(2024)Oquab, Darcet, Moutakanni, Vo, Szafraniec, Khalidov, Fernandez, HAZIZA, Massa, El-Nouby, Assran, Ballas, Galuba, Howes, Huang, Li, Misra, Rabbat, Sharma, Synnaeve, Xu, Jegou, Mairal, Labatut, Joulin, and Bojanowski]{oquab2024dinov}
Maxime Oquab, Timoth{\'e}e Darcet, Th{\'e}o Moutakanni, Huy~V. Vo, Marc Szafraniec, Vasil Khalidov, Pierre Fernandez, Daniel HAZIZA, Francisco Massa, Alaaeldin El-Nouby, Mido Assran, Nicolas Ballas, Wojciech Galuba, Russell Howes, Po-Yao Huang, Shang-Wen Li, Ishan Misra, Michael Rabbat, Vasu Sharma, Gabriel Synnaeve, Hu Xu, Herve Jegou, Julien Mairal, Patrick Labatut, Armand Joulin, and Piotr Bojanowski.
\newblock {DINO}v2: Learning robust visual features without supervision.
\newblock \emph{Transactions on Machine Learning Research}, 2024.
\newblock Featured Certification.

\bibitem[Ranftl et~al.(2021)Ranftl, Bochkovskiy, and Koltun]{Ranftl2021}
Ren\'{e} Ranftl, Alexey Bochkovskiy, and Vladlen Koltun.
\newblock Vision transformers for dense prediction.
\newblock \emph{ICCV}, 2021.

\bibitem[Ranftl et~al.(2022)Ranftl, Lasinger, Hafner, Schindler, and Koltun]{Ranftl2022}
Ren\'{e} Ranftl, Katrin Lasinger, David Hafner, Konrad Schindler, and Vladlen Koltun.
\newblock Towards robust monocular depth estimation: Mixing datasets for zero-shot cross-dataset transfer.
\newblock \emph{IEEE Transactions on Pattern Analysis and Machine Intelligence}, 44\penalty0 (3), 2022.

\bibitem[Saxena et~al.(2009)Saxena, Sun, and Ng]{4531745}
Ashutosh Saxena, Min Sun, and Andrew~Y. Ng.
\newblock Make3d: Learning 3d scene structure from a single still image.
\newblock \emph{IEEE Transactions on Pattern Analysis and Machine Intelligence}, 31\penalty0 (5):\penalty0 824--840, 2009.

\bibitem[Saxena et~al.(2023)Saxena, Kar, Norouzi, and Fleet]{Saxena2023}
Saurabh Saxena, Abhishek Kar, Mohammad Norouzi, and David~J Fleet.
\newblock Monocular depth estimation using diffusion models.
\newblock \emph{arXiv preprint arXiv:2302.14816}, 2023.

\bibitem[Scheerlinck et~al.(2019)Scheerlinck, Rebecq, Stoffregen, Barnes, Mahony, and Scaramuzza]{DAVIS_SCARAMUZZA}
Cedric Scheerlinck, Henri Rebecq, Timo Stoffregen, Nick Barnes, Robert Mahony, and Davide Scaramuzza.
\newblock {CED:} color event camera dataset.
\newblock In \emph{{IEEE} Conf. Comput. Vis. Pattern Recog. Workshops {(CVPRW)}}, 2019.

\bibitem[Shao et~al.(2024)Shao, Yang, Zhou, Zhang, Shen, Poggi, and Liao]{shao2024learning}
Jiahao Shao, Yuanbo Yang, Hongyu Zhou, Youmin Zhang, Yujun Shen, Matteo Poggi, and Yiyi Liao.
\newblock Learning temporally consistent video depth from video diffusion priors.
\newblock \emph{arXiv preprint arXiv:2406.01493}, 2024.

\bibitem[Shi et~al.(2015)Shi, Chen, Wang, Yeung, Wong, and Woo]{shi2015convolutional}
Xingjian Shi, Zhourong Chen, Hao Wang, Dit-Yan Yeung, Wai-Kin Wong, and Wang-chun Woo.
\newblock Convolutional lstm network: A machine learning approach for precipitation nowcasting.
\newblock \emph{Advances in neural information processing systems}, 28, 2015.

\bibitem[Taverni et~al.(2018)Taverni, Paul~Moeys, Li, Cavaco, Motsnyi, San Segundo~Bello, and Delbruck]{DAVIS_CAMERA}
Gemma Taverni, Diederik Paul~Moeys, Chenghan Li, Celso Cavaco, Vasyl Motsnyi, David San Segundo~Bello, and Tobi Delbruck.
\newblock Front and back illuminated dynamic and active pixel vision sensors comparison.
\newblock \emph{IEEE Transactions on Circuits and Systems II: Express Briefs}, 65\penalty0 (5):\penalty0 677--681, 2018.

\bibitem[Yang et~al.(2024{\natexlab{a}})Yang, Kang, Huang, Xu, Feng, and Zhao]{depthanything}
Lihe Yang, Bingyi Kang, Zilong Huang, Xiaogang Xu, Jiashi Feng, and Hengshuang Zhao.
\newblock Depth anything: Unleashing the power of large-scale unlabeled data.
\newblock In \emph{CVPR}, 2024{\natexlab{a}}.

\bibitem[Yang et~al.(2024{\natexlab{b}})Yang, Kang, Huang, Zhao, Xu, Feng, and Zhao]{depth_anything_v2}
Lihe Yang, Bingyi Kang, Zilong Huang, Zhen Zhao, Xiaogang Xu, Jiashi Feng, and Hengshuang Zhao.
\newblock Depth anything v2.
\newblock \emph{arXiv:2406.09414}, 2024{\natexlab{b}}.

\bibitem[Yin et~al.(2020)Yin, Wang, Shen, Liu, Tian, Xu, Sun, and Renyin]{Yin2020}
Wei Yin, Xinlong Wang, Chunhua Shen, Yifan Liu, Zhi Tian, Songcen Xu, Changming Sun, and Dou Renyin.
\newblock Diversedepth: Affine-invariant depth prediction using diverse data.
\newblock \emph{arXiv preprint arXiv:2002.00569}, 2020.

\bibitem[Yin et~al.(2023)Yin, Zhang, Chen, Cai, Yu, Wang, Chen, and Shen]{Yin2023}
Wei Yin, Chi Zhang, Hao Chen, Zhipeng Cai, Gang Yu, Kaixuan Wang, Xiaozhi Chen, and Chunhua Shen.
\newblock {Metric3D}: Towards zero-shot metric 3d prediction from a single image.
\newblock In \emph{ICCV}, 2023.

\bibitem[Zhao et~al.(2022)Zhao, Zhang, Poggi, Tosi, Guo, Zhu, Huang, Tang, and Mattoccia]{zhao2022monovit}
Chaoqiang Zhao, Youmin Zhang, Matteo Poggi, Fabio Tosi, Xianda Guo, Zheng Zhu, Guan Huang, Yang Tang, and Stefano Mattoccia.
\newblock Monovit: Self-supervised monocular depth estimation with a vision transformer.
\newblock In \emph{2022 international conference on 3D vision (3DV)}, pages 668--678. IEEE, 2022.

\bibitem[Zhou et~al.(2017)Zhou, Brown, Snavely, and Lowe]{zhou2017unsupervised}
Tinghui Zhou, Matthew Brown, Noah Snavely, and David~G. Lowe.
\newblock Unsupervised learning of depth and ego-motion from video.
\newblock In \emph{2017 IEEE Conference on Computer Vision and Pattern Recognition (CVPR)}, pages 6612--6619, 2017.

\bibitem[Zhu et~al.(2018{\natexlab{a}})Zhu, Thakur, Özaslan, Pfrommer, Kumar, and Daniilidis]{mvsec}
Alex~Zihao Zhu, Dinesh Thakur, Tolga Özaslan, Bernd Pfrommer, Vijay Kumar, and Kostas Daniilidis.
\newblock The multivehicle stereo event camera dataset: An event camera dataset for 3d perception.
\newblock \emph{IEEE Robotics and Automation Letters}, 3\penalty0 (3):\penalty0 2032--2039, 2018{\natexlab{a}}.

\bibitem[Zhu et~al.(2018{\natexlab{b}})Zhu, Yuan, Chaney, and Daniilidis]{zhu2019unsupervised}
Alex~Zihao Zhu, Liangzhe Yuan, Kenneth Chaney, and Kostas Daniilidis.
\newblock Unsupervised event-based learning of optical flow, depth, and egomotion.
\newblock \emph{CoRR}, abs/1812.08156, 2018{\natexlab{b}}.

\bibitem[Zhu et~al.(2024)Zhu, Liu, Jiang, Wen, Zhang, Li, and Liu]{zhu2024selfsupervisedeventbasedmonoculardepth}
Junyu Zhu, Lina Liu, Bofeng Jiang, Feng Wen, Hongbo Zhang, Wanlong Li, and Yong Liu.
\newblock Self-supervised event-based monocular depth estimation using cross-modal consistency, 2024.

\end{thebibliography}
}

\end{document}